\documentclass{article}
\usepackage{ijcai22}

\usepackage{times}
\usepackage{soul}
\usepackage{url}
\usepackage[hidelinks]{hyperref}
\usepackage[utf8]{inputenc}
\usepackage[small]{caption}
\usepackage{graphicx}
\usepackage{amsmath}
\usepackage{amsthm}
\usepackage{booktabs}
\usepackage{algorithm}
\usepackage{algorithmic}
\usepackage{amssymb}
\usepackage{paralist,bm}
\renewcommand{\vec}[1]{\bm{#1}}
\newcommand{\mat}[1]{\bm{#1}}

\def\eg{\emph{e.g.~}} 
\def\ie{\emph{i.e.~}}

\def\etal{\emph{et al}.~}

\usepackage{xcolor}

\title{BiCo-Net: Regress Globally, Match Locally for Robust 6D Pose Estimation}

\author{
Zelin Xu$^1$
\and
Yichen Zhang$^1$ \and
Ke Chen$^{1,2,}$\footnote{Corresponding authors}
\And
Kui Jia$^{1,2,*}$
\affiliations
$^1$South China University of Technology
\affiliations
$^2$Peng Cheng Laboratory 
\emails
\{eexuzelin, eezyc\}@mail.scut.edu.cn,
\{chenk, kuijia\}@scut.edu.cn
}

\begin{document}

\maketitle

\begin{abstract}
    The challenges of learning a robust 6D pose function lie in 1) severe occlusion and 2) systematic noises in depth images. Inspired by the success of point-pair features, the goal of this paper is to recover the 6D pose of an object instance segmented from RGB-D images by locally matching pairs of oriented points between the model and camera space. To this end, we propose a novel Bi-directional Correspondence Mapping Network (BiCo-Net) to first generate point clouds guided by a typical pose regression, which can thus incorporate pose-sensitive information to optimize generation of local coordinates and their normal vectors. As pose predictions via geometric computation only rely on one single pair of local oriented points, our BiCo-Net can achieve robustness against sparse and occluded point clouds. An ensemble of redundant pose predictions from locally matching and direct pose regression further refines final pose output against noisy observations. Experimental results on three popularly benchmarking datasets can verify that our method can achieve state-of-the-art performance, especially for the more challenging severe occluded scenes. Source codes are available at \url{https://github.com/Gorilla-Lab-SCUT/BiCo-Net}.
\end{abstract}

\section{Introduction}\label{sec:intro}

The problem of 6 Degree-of-Freedom pose estimation (simply put, 6D pose estimation) aims to predict the orientation and location of one detected object instance in 3D space from a canonical model via recovering a rigid SE(3) transformation from the object space to the camera space.  
Such a problem has been widely encountered in the fields of engineering such as robotic grasping and autonomous driving. 
A large number of deep algorithms including regression based \cite{xiang2017posecnn,park2019pix2pose} and keypoint-based \cite{tekin2018real,peng2019pvnet} only rely on extracting texture information from RGB images, which are sensitive to objects with poor textures. 
Alternatively, with the development and a wide application of depth sensors, exploring 6D pose estimation on RGB-D images becomes popular in recent years, where depth images can provide complementary geometry information to RGB images.

\begin{figure}[t]
\centering \includegraphics[width=0.95\linewidth]{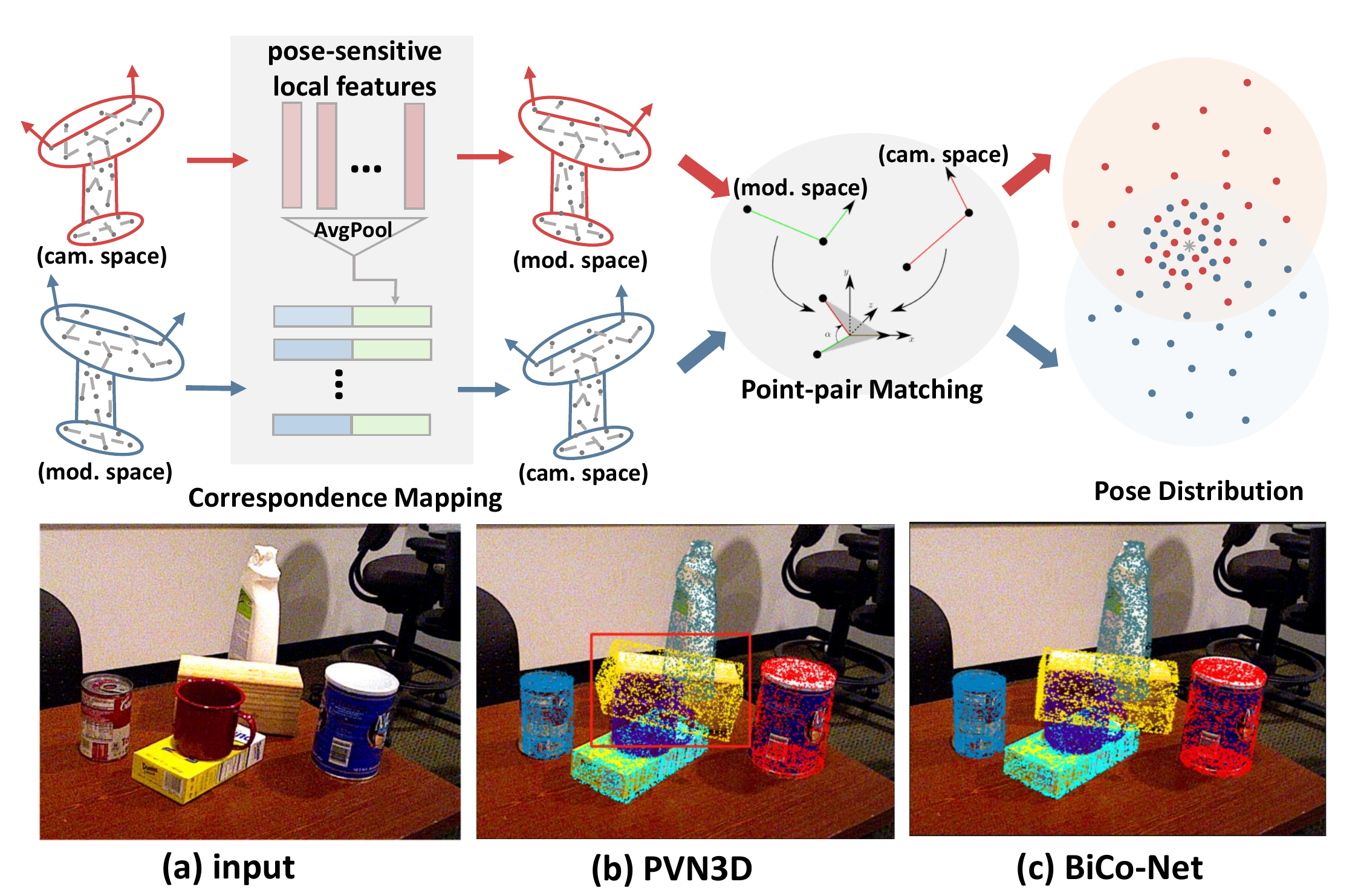}
\caption{Top: Visualization of our BiCo-Net consisting of deep correspondence mapping, pose computation via point pair matching, and an ensemble of pose predictions. Bottom: Comparative results of existing PVN3D and our BiCo-Net with an example from the YCB-Video benchmark, where the red rectangle highlights a failure of 6D pose estimation by the PVN3D due to inter-object occlusion.
}\vspace{-0.5cm}
\label{fig:intro}
\end{figure}  
  
The pioneering works \cite{xiang2017posecnn,li2018unified} on RGB-D images are in a two-stage structure: deep pose estimation and post-processing refinement with iterative closest point (ICP) \cite{besl1992method}, which leads to less efficiency during inference.
Residual learning based refinement in \cite{wang2019densefusion} is proposed to improve efficiency with orders of magnitude faster than the ICP, which can ensure real-time inference.
From a practical perspective, there remain two compounded challenges for learning a robust 6D pose function to address the problem: 1) severe occlusion and 2) unavoidable systematic noises during depth imaging \cite{barron2013intrinsic}. 

Most of the existing methods \cite{wang2019densefusion,He2021FFB6DAF} concern on coping with the former by improving feature encoding on integrating textural and shape features into discriminative representation, while very few work pays attention to the latter.
In \cite{zhou2021pr}, a point refinement network is the first attempt to explicitly polish point clouds via completion and denoising, whose features are combined with those encoded from raw point clouds and RGB images for better multi-modal feature fusion to regress final poses.
However, performance gain of PR-GCN in \cite{zhou2021pr} over the pose regression baseline can be sensitive to point cloud generation, which itself remains an active and challenging task especially under incomplete and noisy observations. 

We argue that \textit{encoding pose-sensitive local features} and \textit{modeling a statistical distribution of pose inliers} are two key factors for accurate and robust performance in 6D pose estimation. 
On one hand, pose estimation dependent on local texture and geometry can perform stably when missing a part of object regions, which is thus robust against occlusion, but those local features are sensitive to the quality (\eg noises and resolution) of the acquired data. 
On the other hand, the distribution of pose predictions on local features can be explored to alleviate negative effects of systematic noises in depth imaging.
In this paper, we propose a novel deep model for 6D pose estimation on RGB-D images, \ie a Bi-directional Correspondence Mapping Network (BiCo-Net) as shown in Figure \ref{fig:intro}, by simultaneously addressing both challenges in a unified and implicit manner.

Given the point cloud (generated from the depth image) and the RGB image of an object under the observation pose, the feature output of DenseFusion feature encoder \cite{wang2019densefusion} is decoded to generate the corresponding oriented points in the canonical space.
To effectively exploit object model priors as a reference, a clean and complete model point cloud under the observed pose is similarly produced in an encoder-decoder learning style from the corresponding oriented points under the canonical pose, which is in an opposite direction of the aforementioned point cloud generation.
Inspired by the success of point-pair features \cite{drost2010model}, a set of oriented point pairs from the input (\eg the camera space) can be randomly sampled to produce accurate pose predictions by matching with those corresponding ones in the output space (\eg the model space).
As pose prediction relies only on one pair of local oriented points, such a characteristic can encourage robustness against sparse and occluded point clouds. 
Moreover, owing to revealing a global distribution of rigid pose transformation favored by multiple point pairs, selecting and combining pose predictions of the bi-directional correspondence mapping can alleviate negative effects of outliers in noisy point clouds.
For imposing pose-sensitive information, features of the bi-directional point cloud generation can be regularized by a typical pose regression as \cite{wang2019densefusion}.

In general, the whole network consists of an ensemble of two parts: direct pose regression on  
a concatenation of global features and point-wise features as \cite{wang2019densefusion} and pose computation via  
locally matching of oriented point pairs in-between canonical and observed poses, which can further refine pose predictions. 
Extensive experiments on three popular benchmarks, \ie YCB-Video \cite{xiang2017posecnn}, LineMOD \cite{hinterstoisser2011multimodal} and Occlusion LineMOD \cite{brachmann2014learning}, can verify superior performance of the proposed BiCo-Net to the state-of-the-art methods, especially for severe occluded scenes. 
Main contributions of this paper lie as follows:
\begin{itemize}
\item This paper proposes a novel 6D pose estimation method -- the BiCo-Net based on locally matching oriented point pairs between the model and camera space and direct pose regression. 
\item The proposed BiCo-Net is implicitly robust against occlusion and sparse point distribution owing to exploiting the pose-sensitive characteristic of each single pair of oriented points under different poses.    
\item Negative effects of outliers in noisy depth images can be mitigated via selection and an ensemble of redundant pose predictions.
\item Our BiCo-Net can gain state-of-the-art performance on three benchmarks, especially on the more challenging Occlusion LineMOD dataset. 
\end{itemize}

\section{Related Works}

\paragraph{Keypoint-based 6D Pose Estimation.}
A typical keypoint-based pose estimation algorithm is designed in a two-stage pipeline: first localizing 2D projection of pre-defined key points in 3D space, and pose predictions can be generated via 2D-to-3D key point correspondence with a PnP \cite{Lepetit2008EPnPAA}. 
Existing methods can be categorized into two groups: object detection based \cite{rad2017bb8,tekin2018real} and dense heatmap based \cite{oberweger2018making,pavlakos20176}.
The former performs well on localizing sparse key points of object foreground, but are sensitive to occlusion \cite{oberweger2018making}. 
The latter group of methods are more robust to inter-object occlusion in a dense correspondence mapping style.
PVNet \cite{peng2019pvnet} is proposed to detect 2D keypoints via voting on pixel-wise predictions of the directional vector that points to keypoints and is robust to truncation and occlusion.
PVN3D \cite{he2020pvn3d} extends the 2D keypoints into 3D space by building 3D-3D correspondence and then uses the least-squares fitting \cite{arun1987least} to generate the pose prediction. 
Learning of directional vectors pointing to 3D keypoints under camera space suffers from the variation of object pose and the vast 3D search space.
In contrast, the bi-directional correspondence mapping in our method makes point-wise predictions on 3D oriented points between the model and camera space directly to build up dense correspondence, instead of their projection in 2D images or the directional vector pointing to 3D keypoints, which enhances local feature discrimination via direct regression based regularization in a pose-sensitive manner. 
\begin{figure*}[t]
\centering \includegraphics[width=0.90\linewidth]{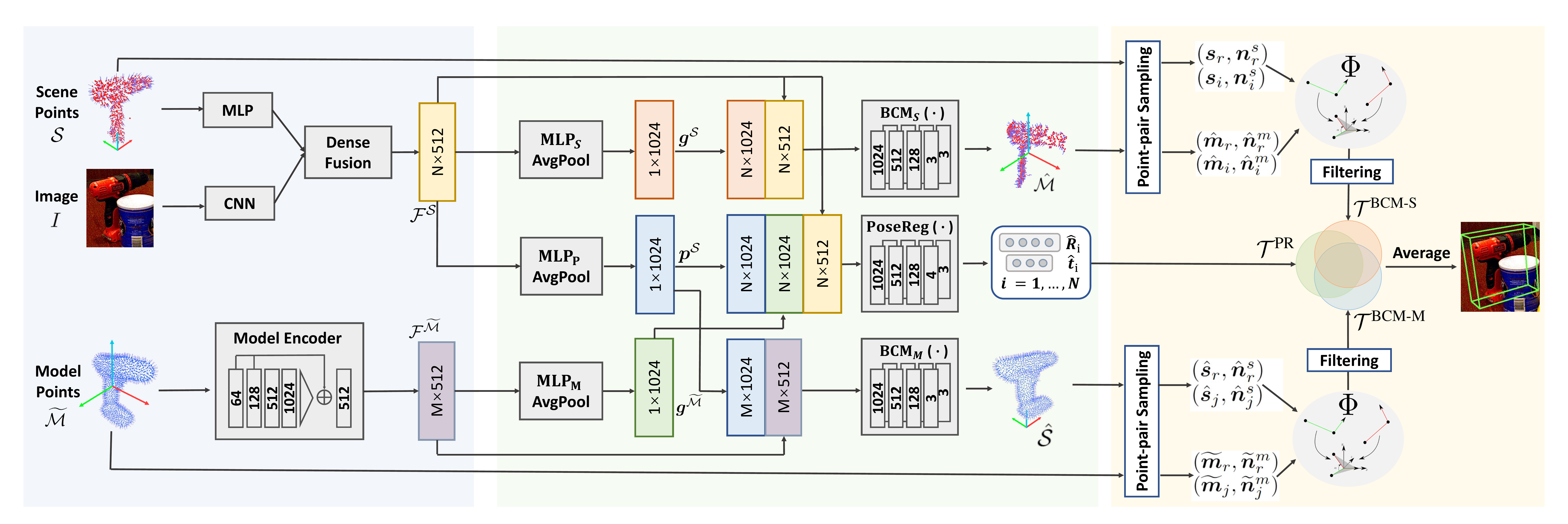}
\caption{Pipeline of the proposed BiCo-Net.
}\vspace{-0.5cm}
\label{fig:our_method}
\end{figure*}

\paragraph{Dense Regression-based 6D Pose Estimation.}
An alternative group of algorithms to cope with occlusion is to produce dense pose predictions for each pixel or local patches with hand-crafted features \cite{liebelt2008independent,sun2010depth}, CNN patch-based feature encoding \cite{doumanoglou2016recovering,kehl2016deep} and CNN pixel-based feature encoding \cite{wang2019densefusion,zhou2021pr}, whose final pose output is selected via a voting scheme.
DenseFusion \cite{wang2019densefusion} fuses RGB and Depth information for each point and uses point-wise features to regress dense poses. 
Zhou \etal proposes PR-GCN \cite{zhou2021pr} to handle incomplete and noisy point clouds in practice via designing a PRN to complete and denoise observed point clouds and use graph convolution to better integrated the RGB and Depth information.
In \cite{wang2019normalized}, normalized object coordinate space is proposed to build up 3D-3D correspondences for each pixel in category level then recover the 6D pose and size by the least-squares fitting.
The proposed BiCo-Net is designed to both regress dense poses from point-wise features and locally match pose-sensitive oriented points in a unified framework. 
The complementary characteristics of dense pose regression and dense correspondence mapping can be fully utilized to gain robust pose predictions. 
Moreover, compared with the least-squares fitting, our point-pair pose computation which uses the correspondence of two oriented points can perform more robust under heavy occlusion.

\section{Methodology}

The problem of 6D pose estimation of an object given the RGB-D image and its canonical CAD model is to estimate their rotation $\mat{R}\in SO(3)$ and translation $\vec{t}\in \mathbb{R}^3$, which can be defined as a rigid transformation $\mat{T}=[\mat{R}|\vec{t}]$ from the model space with respect to the camera space. 
The whole pipeline of our BiCo-Net is illustrated in Figure \ref{fig:our_method}, which can be divided into three modules: feature encoding on segmented object instances (see Sec. \ref{subsec:feat_fusion}) in a blue block, generation of oriented points via Bi-directional Correspondence Mapping (see Sec. \ref{subsec:BCM}) in a green block, and an ensemble of pose predictions by point pair matching and point-wise regression (see Sec. \ref{subsec:point_pair_pose}) in a yellow block.

\subsection{Instance Segmentation and Feature Encoding}\label{subsec:feat_fusion}
Following existing methods \cite{wang2019densefusion,zhou2021pr}, our BiCo-Net first employs an off-the-shelf instance segmentation method for RGB images (\eg Mask RCNN \cite{he2017mask} in our experiments) to segment the object of interests, which produces a cropped image patch $I$ and one scene point cloud under the camera space $\mathcal{S} = \{(\bm{s}_i, \bm{n}^s_i) \in \mathbb{R}^6 \}_{i=1}^{N}$, where $\bm{s}_i \in \mathbb{R}^3$ denotes 3D coordinates converted from the masked depth region and $\bm{n}^s_i \in \mathbb{R}^3$ denotes its normal vectors physically computed by PCA.
Both $I$ and $\mathcal{S}$ are fed into CNN-based and MLP-based feature encoders respectively to extract texture and geometric features from heterogeneous data sources, which are further fused by the DenseFusion module introduced in \cite{wang2019densefusion} to obtain point-wise features $\mathcal{F}^\mathcal{S} = \{\bm{f}^s_i \in \mathbb{R}^{512}\}_{i=1}^{N}$. 
Similarly, for exploiting the model priors, we randomly sample one clean model point cloud $\widetilde{\bm{m}}$ from the canonical CAD model and compute point-wise normal vectors $\widetilde{\bm{n}}^m$ to form $\widetilde{\mathcal{M}} = \{(\widetilde{\bm{m}}_j, \widetilde{\bm{n}}^m_j) \in \mathbb{R}^6 \}_{j=1}^{M}$  as input of an MLP-based feature encoder to generate point-wise features $\mathcal{F}^{\widetilde{\mathcal{M}}} = \{\widetilde{\bm{f}}^m_j \in \mathbb{R}^{512}\}_{j=1}^{M}$.

\subsection{Bi-directional Correspondence Mapping}\label{subsec:BCM}
Given features $\mathcal{F}^\mathcal{S}$ from visual observation $I$ and $\mathcal{S}$, the BCM-scene (BCM-S) in the top row of Figure \ref{fig:our_method} aims to regress the corresponding oriented point cloud under the model space $\mathcal{M} = \{(\bm{m}_i, \bm{n}^m_i) \in \mathbb{R}^6 \}_{i=1}^{N}$, where $\bm{m}_i = \mat{R}^{-1}(\bm{s}_i-\vec{t})$ and $\bm{n}^m_i = \mat{R}^{-1}\bm{n}^s_i$. 
With the point-wise features $\mathcal{F}^\mathcal{S}$ as input, we employ a MLP with {{\{512,1024\} neurons}} and an average pooling (AvgPool) to generate a global feature $\vec{g}^\mathcal{S} = \texttt{AvgPool}(\texttt{MLP}_S(\mathcal{F}^\mathcal{S})) \in \mathbb{R}^{1024}$.
Finally, $\mathcal{\hat{M}} = \texttt{BCM}_S(\mathcal{F}^\mathcal{S}, \vec{g}^\mathcal{S})$ for superior robustness to only using $\mathcal{F}^\mathcal{S}$, where generated points $\mathcal{\hat{M}} = \{(\hat{\bm{m}}_i, \hat{\bm{n}}^m_i) \}_{i=1}^{N} $.
Similarly, the BCM-model (BCM-M) in the bottom row of Figure \ref{fig:our_method} aims to reconstruct a clean point cloud under the camera space $\widetilde{\mathcal{S}} = \{(\widetilde{\bm{s}}_j, \widetilde{\bm{n}}^s_j) \in \mathbb{R}^6 \}_{j=1}^{M}$ from the features under the model space $\mathcal{F}^{\widetilde{\mathcal{M}}}$, where $\widetilde{\bm{s}}_j = \mat{R}\widetilde{\bm{m}}_j+\vec{t}$ and $\widetilde{\bm{n}}^s_j = \mat{R}\widetilde{\bm{n}}^m_j$.
To this end, a global feature $\vec{p}^\mathcal{S} \in \mathbb{R}^{1024}$ aggregated from $\mathcal{F}^\mathcal{S}$ together with $\mathcal{F}^{\widetilde{\mathcal{M}}}$ are learning to regress $\widetilde{\mathcal{S}}$, while $\hat{{\mathcal{S}}} = \texttt{BCM}_M(\mathcal{F}^{\widetilde{\mathcal{M}}}, \vec{p}^\mathcal{S})$ is the generated point cloud under the camera space by the BCM-M branch. 
Moreover, to impose pose sensitive information for generation of point clouds in the BCM-S and BCM-M, a direct point-wise pose regression (PR) in the middle row of Figure \ref{fig:our_method} on $\mathcal{F}^\mathcal{S}$, $\vec{g}^{\widetilde{\mathcal{M}}}$, and $\vec{p}^\mathcal{S}$ to predict $\hat{\mathcal{T}}^\text{PR} = \texttt{PoseReg}(\mathcal{F}^\mathcal{S}, \vec{g}^{\widetilde{\mathcal{M}}}, \vec{p}^\mathcal{S})$, where $\hat{\mathcal{T}}^\text{PR} = \{(\hat{\mat{R}}_i, \hat{\vec{t}}_i)\}_{i=1}^{N}$ and $\vec{g}^{\widetilde{\mathcal{M}}} = \texttt{AvgPool}(\texttt{MLP}_M(\mathcal{F}^{\widetilde{\mathcal{M}}}))$. 

\paragraph{Loss Functions.}
We use the Euclidean distance to supervise both BCM branches as follows:
\begin{equation*}
\begin{split}
L^{\text{BCM-S}}\!=\!\frac{1}{N}\sum_i(||\bm{m}_i\!-\! \hat{\bm{m}}_i||\!+\!\lambda||\bm{n}^m_i\!-\!\hat{\bm{n}}^m_i||),\\
L^{\text{BCM-M}}=\frac{1}{M}\sum_j(||\widetilde{\bm{s}}_j-\hat{{\bm{s}}}_j||+\lambda||\widetilde{\bm{n}}^s_j-\hat{{\bm{n}}}^s_j||),
\end{split}
\end{equation*}
where $\lambda$ is the trade-off parameter between two terms, $(\bm{m}_i, \bm{n}^m_i)$ and $(\widetilde{\bm{s}}_j, \widetilde{\bm{n}}^s_j)$ are ground truth oriented points of the BCM-S and BCM-M branches, while $(\hat{\bm{m}}_i, \hat{\bm{n}}^m_i)$, $(\hat{{\bm{s}}}_j, \hat{{\bm{n}}}^s_j)$ are the generated points.
To ensure the pose consistency between BCM branches and direct regression branch, we replace the ground truth pose in $(\bm{m}_i, \bm{n}^m_i)$, $(\widetilde{\bm{s}}_j, \widetilde{\bm{n}}^s_j)$ with the mean of point-wise predicted pose $[\hat{\mat{R}}_i|\hat{\vec{t}}_i]$ for symmetric objects.
For supervising $\hat{\mathcal{T}}^\text{PR}$ with $\mathcal{T}=[\mat{R}|\vec{t}]$ in the pose regression branch, we use the ADD Loss \cite{xiang2017posecnn} for asymmetric objects and ADD-S Loss for symmetric objects:
\begin{equation*}
L^{\text{PR}}_i\!=\!\begin{cases}
\frac{1}{K}\sum_k{ || (\mat{R}\vec{x}_k+\vec{t})- (\hat{\mat{R}}_i\vec{x}_k+\hat{\vec{t}}_i) || } 
\qquad~~~~~\enspace{\text{if asym.}} \,,\\
\frac{1}{K}\!\sum_k{\!\min_{0<l<K}|| (\mat{R}\vec{x}_k\!+\!\vec{t}) - (\hat{\mat{R}}_i\vec{x}_l\!+\!\hat{\vec{t}}_i) || }
\enspace\text{if sym.} \,;
\end{cases}
\end{equation*}
where $K$ is the number of points sampled from  the surface of the CAD model,
$[\mat{R}|\vec{t}]$ and $[\hat{\mat{R}}_i|\hat{\vec{t}}_i]$ are ground truth and point-wise predicted poses respectively.
The total loss of our BiCo-Net can thus be written as:
\begin{equation*}
L^{\text{Total}}=\frac{1}{N}\sum_iL^{\text{PR}}_i + L^{\text{BCM-S}} + L^{\text{BCM-M}}.
\end{equation*}

\begin{table*}[!thp]
\renewcommand\arraystretch{1.2}
\centering
\caption{Comparison of AUC (\%) and ADD-S \textless{} 2cm (\%) (``\textless{}2cm'' for short) on the YCB-Video dataset. Symmetric objects are highlighted in bold. Comparative methods with the proposed BiCo-Net are PoseCNN+ICP \protect\cite{xiang2017posecnn}, DenseFusion \protect\cite{wang2019densefusion}, G2L-Net \protect\cite{chen2020g2l}, PVN3D \protect\cite{he2020pvn3d} and PR-GCN \protect\cite{zhou2021pr}.}
\label{tab:ycb_result}
\resizebox{0.75\textwidth}{!}{%
\centering
\begin{tabular}{l|cc|cc|cc|cc|cc|cc}
\hline
& \multicolumn{2}{c|}{PoseCNN+ICP} & \multicolumn{2}{c|}{DenseFusion} & \multicolumn{2}{c|}{G2L-Net} & \multicolumn{2}{c|}{PVN3D} & \multicolumn{2}{c|}{PR-GCN} & \multicolumn{2}{c}{BiCo-Net (ours)}
\\ \hline
& AUC & \textless{}2cm & AUC & \textless{}2cm & AUC & \textless{}2cm & AUC & \textless{}2cm & AUC & \textless{}2cm & AUC & \textless{}2cm \\
\hline
002\_master\_chef\_can & 95.8 & \textbf{100.0} & 96.4 & \textbf{100.0} & 94.0 & - & 96.0 & \textbf{100.0} & \textbf{97.1} & \textbf{100.0} & 96.2 & \textbf{100.0} \\ 
003\_cracker\_box & 92.7 & 91.6 & 95.5 & 99.5 & 88.7 & - & 96.1 & \textbf{100.0} & \textbf{97.6} & \textbf{100.0} & 96.6 & \textbf{100.0} \\ 
004\_sugar\_box & 98.2 & \textbf{ 100.0} & 97.5 & \textbf{100.0} & 96.0 & - & 97.4 & \textbf{100.0} &\textbf{98.3} & \textbf{100.0} & 97.8 & \textbf{100.0} \\ 
005\_tomato\_soup\_can & 94.5 & 96.9  & 94.6 & 96.9 & 86.4 & - & \textbf{96.2} & \textbf{98.1} & 95.3 & 97.6 & 95.7 & \textbf{98.1} \\ 
006\_mustard\_bottle & \textbf{98.6} & \textbf{100.0}  & 97.2 & \textbf{100.0} & 95.9 & - & 97.5 & \textbf{100.0} & 97.9 & \textbf{100.0} & 98.0 & \textbf{100.0} \\ 
007\_tuna\_fish\_can & 97.1 & \textbf{100.0} & 96.6 & \textbf{100.0} & 84.1 & - & 96.0 & \textbf{100.0} & \textbf{97.6} & \textbf{100.0} & 96.5 & \textbf{100.0} \\ 
008\_pudding\_box & 97.9 & \textbf{100.0} & 96.5 & \textbf{100.0} & 93.5 & - & 97.1 & \textbf{100.0} & \textbf{98.4} & \textbf{100.0} & 97.5 & \textbf{100.0} \\ 
009\_gelatin\_box & \textbf{98.8} & \textbf{100.0}  & 98.1 & \textbf{100.0} & 96.8 & - & 97.7 & \textbf{100.0} & 96.2 & 94.4 & \textbf{98.8} & \textbf{100.0} \\ 
010\_potted\_meat\_can & 92.7 & 93.6 & 91.3 & 93.1 & 86.2 & - & 93.3 & 94.6 & \textbf{96.6} & \textbf{99.1} & 93.0 & 94.5 \\ 
011\_banana & 97.1 & 99.7 & 96.6 & \textbf{100.0} & 96.3 & - & 96.6 & \textbf{100.0} & \textbf{98.5} & \textbf{100.0} & 97.1 & \textbf{100.0} \\ 
019\_pitcher\_base & 97.8 & \textbf{100.0} & 97.1 & \textbf{100.0} & 91.8 & - & 97.4 & \textbf{100.0} & \textbf{98.1} & \textbf{100.0} & 97.6 & \textbf{100.0} \\ 
021\_bleach\_cleanser & 96.9 & 99.4 & 95.8 & \textbf{100.0} & 92.0 & - & 96.0 & \textbf{100.0} & \textbf{97.9} & \textbf{100.0} & 96.6 & \textbf{100.0} \\ 
\textbf{024\_bowl} & 81.0 & 54.9 & 88.2 & \textbf{98.8} & 86.7 & - & 90.2 & 80.5 & 90.3 & 96.6 & \textbf{96.7} & \textbf{100.0} \\
025\_mug & 95.0 & 99.8 & 97.1 & \textbf{100.0} & 95.4 & - & 97.6 & \textbf{100.0} & \textbf{98.1} & \textbf{100.0} & 97.0 & \textbf{100.0} \\ 
035\_power\_drill & \textbf{98.2} & 99.6 & 96.0 & 98.7 & 95.2 & - & 96.7 & \textbf{100.0} & 98.1 & \textbf{100.0} & 97.0 & 99.9 \\ 
\textbf{036\_wood\_block} & 87.6 & 80.2 & 89.7 & 94.6 & 86.2 & - & 90.4 & 93.8 & \textbf{96.0} & \textbf{100.0} & 92.1 & 90.1 \\ 
037\_scissors & 91.7 & 95.6 & 95.2 & \textbf{100.0} & 83.8 & - & \textbf{96.7} & \textbf{100.0} & \textbf{96.7} & \textbf{100.0} & 92.2 & 99.5 \\ 
040\_large\_marker & 97.2 & 99.7 & 97.5 & \textbf{100.0} & 96.8 & - & 96.7 & 99.8 & \textbf{97.9} & \textbf{100.0} & 97.4 & \textbf{100.0} \\ 
\textbf{051\_large\_clamp} & 75.2 & 74.9 & 72.9 & 79.2 & 94.4 & - & 93.6 & 93.6 & 87.5 & 93.3 & \textbf{94.7} & \textbf{98.3} \\ 
\textbf{052\_extra\_large\_clamp} & 64.4 & 48.8 & 69.8 & 76.3 & \textbf{92.3} & - & 88.4 & 83.6 & 79.7 & 84.6 & 88.2 & \textbf{90.2} \\ 
\textbf{061\_foam\_brick} & 97.2 & \textbf{100.0} & 92.5 & \textbf{100.0} & 94.7 & - & 96.8 & \textbf{100.0} & \textbf{97.8} & \textbf{100.0} & 97.2 & \textbf{100.0} \\ 
\hline
ALL & 93.0 & 93.2 & 93.1 & 96.8 & 92.4 & - & 95.5 & 97.6 & 95.8 & 98.5 & \textbf{96.0} & \textbf{98.8} \\ \hline
\end{tabular}%
}
\vspace{-0.2cm}
\end{table*}

\begin{table*}[t]
\begin{center}
\caption{Comparative evaluation of 6D pose estimation in terms of ADD(-S) (\%) on the LineMOD dataset. Objects in bold are symmetric.}
\label{tab:limo_result}
\resizebox{0.9\textwidth}{!}
		{
\begin{tabular}{l|ccccccccccccc|c}
\hline
Method & ape & ben. & cam & can & cat & drill. & duck & \textbf{egg.} & \textbf{glue} & hole. & iron & lamp & phone & MEAN \\ \hline
Implict+ICP \cite{sundermeyer2018implicit} & 20.6 & 64.3 & 63.2 & 76.1 & 72.0 & 41.6 & 32.4 & 98.6 & 96.4 & 49.9 & 63.1 & 91.7 & 71.0 & 64.7 \\
SSD6D+ICP \cite{kehl2017ssd} & 65.0 & 80.0 & 78.0 & 86.0 & 70.0 & 73.0 & 66.0 & \textbf{100.0} & \textbf{100.0} & 49.0 & 78.0 & 73.0 & 79.0 & 79.0 \\
PointFusion \cite{xu2018pointfusion} & 70.4 & 80.7 & 60.8 & 61.1 & 79.1 & 47.3 & 63.0 & 99.9 & 99.3 & 71.8 & 83.2 & 62.3 & 78.8 & 73.7 \\
DenseFusion \cite{wang2019densefusion} & 79.5 & 84.2 & 76.5 & 86.6 & 88.8 & 77.7 & 76.3 & 99.9 & 99.4 & 79.0 & 92.1 & 92.3 & 88.0 & 86.2 \\
DenseFusion (Iter.) \cite{wang2019densefusion} & 92.3 & 93.2 & 94.4 & 93.1 & 96.5 & 87.0 & 92.3 & 99.8 & \textbf{100.0} & 92.1 & 97.0 & 95.3 & 92.8 & 94.3 \\
G2L-Net \cite{chen2020g2l} & 96.8 & 96.1 & 98.2 & 98.0 & 99.2 & \textbf{99.8} & 97.7 & \textbf{100.0} & \textbf{100.0} & 99.0 & 99.3 & 99.5 & 98.9 & 98.7 \\
PR-GCN \cite{zhou2021pr} & \textbf{97.6} & \textbf{99.2} & 99.4 & 98.4 & 98.7 & 98.8 & \textbf{98.9} & 99.9 & \textbf{100.0} & \textbf{99.4} & 98.5 & 99.2 & 98.4 & 98.9 \\
\hline
BiCo-Net (Ours) & 97.3 & 98.8 & \textbf{99.6} & \textbf{99.3} & \textbf{100.0} & 98.9 & 98.7 & 99.8 & 99.8 & 99.2 & \textbf{100.0} & \textbf{99.7} & \textbf{99.2} & \textbf{99.3} \\
\hline
\end{tabular}\label{tab.limo}
}
\end{center}\vspace{-0.3cm}
\end{table*}

\subsection{Point Pair Matching and Prediction Ensemble}\label{subsec:point_pair_pose}
With generated point clouds $\hat{\mathcal{M}}$ and $\hat{\mathcal{S}}$ under the model and camera space respectively, our goal is to learn a rigid transformation $\mat{T}$ or $\mat{T}^{-1}$ between camera and model space. 
Encouraged by Point-pair feature (PPF) \cite{drost2010model} to describe object poses by matching local features generated from oriented point pairs, any pair of oriented points in $\mathcal{S}$ and its corresponding ones in $\hat{\mathcal{M}}$ can determine object pose, while similar situation is observed for point pairs between $\widetilde{\mathcal{M}}$ and $\hat{\mathcal{S}}$.
Specifically, object pose can be readily obtained by randomly sampling a point-pair $(\bm{s}_r, \bm{s}_i)$ from scene points $\mathcal{S}$ or $(\widetilde{\bm{m}}_r, \widetilde{\bm{m}}_j)$ from model points $\mathcal{\widetilde{\mathcal{M}}}$ and then matching it with the corresponding pair $(\hat{\bm{m}}_r, \hat{\bm{m}}_i)$ or $(\hat{{\bm{s}}}_r, \hat{{\bm{s}}}_j)$ generated by our BCM branches. 
The transformation from $(\hat{\bm{m}}_r, \hat{\bm{m}}_i)$ to $(\bm{s}_r, \bm{s}_i)$ is defined as the following \cite{drost2010model}:
\begin{equation*}
\bm{s}_i = \mat{T}^{-1}_{\bm{s}_r\rightarrow x}\mat{R}_x(\alpha)\mat{T}_{\hat{\bm{m}}_r \rightarrow x}\hat{\bm{m}}_i,
\end{equation*}
where $\mat{T}_{\hat{\bm{m}}_r \rightarrow x}$ denotes a transformation that translates $\hat{\bm{m}}_r$ into the origin and rotates $\hat{\bm{n}}^m_r$ on to the $x$-axis, and the same definition of $\mat{T}_{\bm{s}_r \rightarrow x}$ for transforming $(\bm{s}_r, \bm{n}^s_r)$. 
When $\hat{\bm{m}}_r$ and $\bm{s}_r$ are aligned to the $x$-axis, there is a $\alpha$ angle difference about the $x$-axis between $\mat{T}_{\bm{s}_r\rightarrow x}\bm{s}_i$ and $\mat{T}_{\hat{\bm{m}}_r \rightarrow x}\hat{\bm{m}}_i$, which encourages to use $\mat{R}_x(\alpha)$, a rotation with respect to the $x$-axis, to align $\mat{T}_{\bm{s}_r\rightarrow x}\bm{s}_i$ and $\mat{T}_{\hat{\bm{m}}_r \rightarrow x}\hat{\bm{m}}_i$. 
As a result, the rigid transformation $\Phi: (\bm{s}_r, \bm{s}_i, \hat{\bm{m}}_r, \hat{\bm{m}}_i) \rightarrow \mat{T}$  can be written as:
\begin{equation*}
\Phi(\bm{s}_r, \bm{s}_i, \hat{\bm{m}}_r, \hat{\bm{m}}_i) = \mat{T}^{-1}_{\bm{s}_r\rightarrow x}\mat{R}_x(\alpha)\mat{T}_{\hat{\bm{m}}_r \rightarrow x}.
\end{equation*}
Note that, 6D object pose by locally matching a pair of oriented points under the camera and model space can be readily computed for desirable real-time inference.

As our method only relies on each single point pair to estimate 6D object pose, it allows sparse and imbalanced point distributions, which is thus able to achieve good performance for severe occlusion.  
For alleviating noises in point clouds and increasing inference speed, we use the FPS algorithm to downsample $\mathcal{S}$ and $\widetilde{\mathcal{M}}$ to a subset of $Z$ points, which are then to generate $Z^2$ point pairs to compute pair-wise pose candidates $\mat{T}_z = [\mat{R}_z|\vec{t}_z]$ for the BCM-S and BCM-M respectively. 
For avoiding unreliable pose candidates from point-pairs constructed by neighboring points, pose predictions will be filtered out with the following error measure:
\begin{equation*}
\mathcal{E}(\mat{T}_z) = \frac{1}{N}\sum_i||(\mat{R}^{-1}_z(\bm{s}_i - \vec{t}_z)) - \hat{\bm{m}}_i ||,
\end{equation*}
and preserving the top 10\% of candidates as the pose prediction sets $\mathcal{T}^\text{BCM-S}$ and $\mathcal{T}^\text{BCM-M}$ of these two branches.

\paragraph{An Ensemble of Pose Predictions.}
As mentioned in Sec. \ref{sec:intro}, we can obtain three sets of pose predictions (\ie $\mathcal{T}^\text{PR}$ from direct pose regression; $\mathcal{T}^\text{BCM-S}$ and $\mathcal{T}^\text{BCM-M}$ via locally matching with point pairs), from three branches of the BiCo-Net.
For achieving superior robustness via using the complementary information of three sets, we consider applying the average pose of $\mathcal{T}^\text{PR}\cup\mathcal{T}^\text{BCM-S}\cup\mathcal{T}^\text{BCM-M}$ as the final pose output.

\section{Experiments}

\subsection{Datasets and Settings}
\paragraph{Datasets.}
To evaluate our BiCo-Net comprehensively, experiments are conducted on three popular benchmarks -- the YCB-Video dataset \cite{xiang2017posecnn}, the LineMOD \cite{hinterstoisser2011multimodal}, and the more challenging Occlusion LineMOD \cite{brachmann2014learning}.
The YCB-Video dataset has 92 videos in total, each of which shows a subset of 21 objects with varying textures and sizes under cluttered indoor environment. 
Following existing works \cite{wang2019densefusion,zhou2021pr}, we adopt 16,189 frames from 80 videos with an additional 80,000 synthetic images provided by \cite{xiang2017posecnn} for training and extract 2949 key frames from the remaining 12 videos for testing.
The LineMOD contains 15,783 images belonging to 13 low-textural objects placed under different cluttered environments.
We use the standard training/testing split as \cite{xiang2017posecnn,wang2019densefusion}.
The Occlusion LineMOD provides 6D pose labels of 8 objects selected from the LineMOD and includes 1214 images with multiple heavily occluded objects, which is made more challenging.

\paragraph{Performance Metrics.}
Following \cite{xiang2017posecnn,wang2019densefusion}, we adopt the average distance (ADD) \cite{xiang2017posecnn} and ADD-Symmetric (ADD-S) as performance metrics. 
6D pose predictions are considered to be correct if the ADD/ADD-S is smaller than a predefined threshold.
For the YCB-Video dataset, we vary from 0 to 10cm to plot an accuracy-threshold curve and report the area under the curve (AUC). We also report the result of ADD-S \textless 2cm as \cite{wang2019densefusion,zhou2021pr}.
For the LineMOD and the Occlusion LineMOD, following \cite{he2020pvn3d,zhou2021pr}, we use ADD-S for symmetric objects (\ie eggbox and glue) and ADD for the remaining objects having an asymmetric geometry while taking 10\% of the diameter as the threshold.

\subsection{Implementation Details}
The numbers of scene/model points, \ie, $N$/$M$, are set to 1000/1000. 
In point-pair pose computation, we downsample the scene points and model points to $Z=100$ points by the FPS which thus generates $Z^2 = 10,000$ pose candidates from point pairs. 
The hyper-parameter $\lambda$ in the losses of BCM-S and BCM-M branches is empirically set to 0.05.
We use the Adam optimizer with a $10^{-4}$ learning rate to train our model for 50 epochs, and the learning rate decays 0.3 per 10 epochs.

\begin{table}[t]
\centering
\caption{
Comparison of ADD(-S) (\%) on the Occlusion LineMOD. Symmetric objects are marked in bold. 
Competing methods with our BiCo-Net are PoseCNN \protect\cite{xiang2017posecnn}, Pix2pose \protect\cite{park2019pix2pose}, PVNet \protect\cite{peng2019pvnet}, HybridPose \protect\cite{song2020hybridpose}, PVN3D \protect\cite{he2020pvn3d} and PR-GCN \protect\cite{zhou2021pr}. {}}
\label{tab:occ_limo_result}
 \resizebox{0.5\textwidth}{!}{
\begin{tabular}{l|cccccc|c}
\hline
Method & PoseCNN & Pix2pose & PVNet & HybridPose & PVN3D & PR-GCN  & BiCo-Net\\
\hline
ape & 9.6 & 22.0 & 15.8 & 20.9 & 33.9 & 40.2 & \textbf{55.6}\\
can  & 45.2 & 44.7 & 63.3 & 75.3 & \textbf{88.6} & 76.2 & 83.2\\
cat & 0.9 & 22.7 & 16.7  & 24.9 & 39.1 & \textbf{57.0} & 47.3 \\
drill. & 41.4  & 44.7 & 65.7 & 70.2 & 78.4 & \textbf{82.3} & 69.9 \\ 
duck & 19.6 & 15.0 & 25.2 & 27.9 & 41.9 & 30.0 & \textbf{58.3} \\ 
\textbf{egg.} & 25.9 & 25.2 & 50.2 & 52.4 & \textbf{80.9} & 68.2 & 78.1\\
\textbf{glue} & 39.6 & 32.4 & 49.6 & 53.8 & 68.1 & 67.0 & \textbf{76.9} \\ 
holep. & 22.1 & 49.5 & 39.7 & 54.2 & 74.7 & \textbf{97.2} & 87.2\\
\hline
MEAN & 24.9 & 32.0 & 40.8 & 47.5 & 63.2 & 65.0 & \textbf{69.5}\\ 
\hline
\end{tabular}
}
\vspace{-0.2cm}
\end{table}

\subsection{Comparison with State-of-the-art Methods}
Comparative evaluation of the proposed BiCo-Net and state-of-the-art methods on the YCB-Video, LineMOD and Occlusion LineMOD datasets are showed in Tables \ref{tab:ycb_result}, \ref{tab:limo_result}, and \ref{tab:occ_limo_result}. 
In general, our method can consistently achieve state-of-the-art performance in all benchmarks.
Specifically, on the YCB-Video dataset, our method achieves the best performance on both metrics in Table \ref{tab:ycb_result}, and similar results on the LineMOD in Table \ref{tab:limo_result} can also be observed. 
Compared to moderate improvement on the  YCB-Video and LineMOD, the proposed BiCo-Net can gain accuracy of 69.5\% on the more challenging Occlusion LineMOD, which is significantly superior to the state-of-the-art methods as illustrated in Table \ref{tab:occ_limo_result}. 
Such results can verify the effectiveness of our BiCo-Net for 6D pose estimation on RGB-D images.
In addition, we measure the inference time on average of the proposed method: the forward time of BiCo-Net is 16ms; the point-pair pose computation time is 29ms; the segmentation network takes 30ms. 
As a result, the average time for processing a frame for inference is 75ms with a GTX 1080 Ti GPU, which is comparable to existing methods (\eg 60ms for the DenseFusion and 68ms for the PR-GCN) to meet desirable real-time inference in practical applications.

\subsection{Ablation Studies}

\begin{figure}[t]
\centering \includegraphics[width=0.85\linewidth]{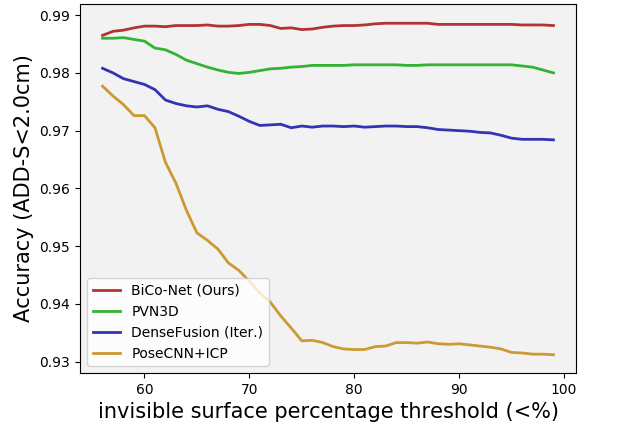}
\caption{Comparative evaluation with different levels of occlusion on the YCB-Video benchmark.}
\label{fig:occ_robust}\vspace{-0.5cm}
\end{figure}

\paragraph{Robustness against Inter-Object Occlusion.}
To evaluate the robustness of our method against occlusion, we follow \cite{wang2019densefusion,he2020pvn3d} to define occlusion level by using the percentage of invisible points on the object surface. 
We compare our BiCo-Net with several methods, including PoseCNN+ICP \cite{xiang2017posecnn}, DenseFusion \cite{wang2019densefusion}, and PVN3D \cite{he2020pvn3d}, on the YCB-Video with accuracy of ADD-S \textless 2cm under different levels of occlusion, whose results are shown in Figure \ref{fig:occ_robust}. 
With the percentage of invisible part increasing, our method can perform stably well when performance of comparative methods decrease or even collapse, which again confirms superior robustness of the proposed BiCo-Net to its competitors  
even with heavily occluded objects.

\begin{table}[h]
\centering
\caption{Effects of  bi-directional correspondence mapping (BCM) branches and an ensemble of pose predictions based on pose regression (Pose Reg). We report the AUC (\%) on YCB-Video (YCB-V) and ADD(-S) (\%) on Occlusion LineMOD (LM-O).}
\label{tab:ablation_bcm}
\resizebox{0.40\textwidth}{!}{
\begin{tabular}{cccc|cc}
\hline
Pose Reg &BCM-S & BCM-M & Ensemble &YCB-V & LM-O \\
\hline
\checkmark & $\times$ & $\times$ & $\times$ & 94.6 & 62.1\\
\hline
\checkmark & \checkmark & $\times$ & $\times$ & 95.3 & 63.7\\
\checkmark & \checkmark & $\times$ & \checkmark & 95.5 & 66.9\\
\hline
\checkmark & $\times$ & \checkmark & $\times$ & 95.4 & 66.1\\
\checkmark & $\times$ & \checkmark & \checkmark & 95.5 & 67.4\\
\hline
\checkmark & \checkmark & \checkmark & $\times$ & 95.7 & 67.6\\
\checkmark & \checkmark & \checkmark & \checkmark & 96.0 & 69.5\\
\hline
\end{tabular}}\vspace{-0.5cm}
\end{table}

\paragraph{Effects of An Ensemble of Filtered Pose Predictions.}
To learn a correspondence mapping using multiple instances, a typical robust scheme is least-square fitting based RANSAC to optimize the hypothesis with the maximum inliers. 
We conduct one experiment in terms of the AUC metrics of the YCB-Video dataset to obtain 79.1\%/89.1\% on the output of the BCM-S/BCM-M branch, which is significantly inferior to the results using point-pair matching in our BiCo-Net, reaching 94.9\%/95.0\% only with $\mathcal{T}^\text{BCM-S}$/$\mathcal{T}^\text{BCM-M}$ (\ie without direct pose regression).
Such a result confirms the rationale of ensembling pose predictions in $\mathcal{T}^\text{BCM-S}$ and $\mathcal{T}^\text{BCM-M}$, owing to the better robustness against outliers in noisy point clouds than the RANSAC. 
Moreover, we take the average of $\mathcal{T}^\text{PR}\cup\mathcal{T}^\text{BCM-S}\cup\mathcal{T}^\text{BCM-M}$ as the final pose prediction of our BiCo-Net in an ensemble manner, whose effectiveness is verified on the YCB-Video and Occlusion LineMOD respectively (See the last row of Table \ref{tab:ablation_bcm}).
Moreover, without direct pose regression, as results of $\mathcal{T}^\text{BCM-S}$/$\mathcal{T}^\text{BCM-M}$ are only 94.9\%/95.0\% on the AUC metrics of the YCB-Video dataset, the direct pose regression branch can benefit generation of point clouds in the BCM branches (\ie 95.7\%/95.6\% for $Z=100$ in Table \ref{tab:ablation_pair}).

\paragraph{Effects of Bidirectional Correspondence Mapping.}
For evaluating the effectiveness of the BCM-S and BCM-M branches of our method, we conduct experiments on the YCB-Video dataset and the Occlusion LineMOD. 
The baseline method takes scene points $\mathcal{S}$ and cropped image patch $I$ as input and only performs point-wise pose regression by $\mathcal{T}^\text{PR} = \texttt{PoseReg}(\mathcal{F}^\mathcal{S}, \vec{p}^\mathcal{S})$. 
The average of $\mathcal{T}^\text{PR}$ of each object instance is utilized as its pose prediction.
As shown in Table \ref{tab:ablation_bcm}, the method introducing the BCM-S branch can improve 0.7\% and 1.6\% on two datasets respectively, indicating that BCM-S effectively improved the discrimination of local feature coding owing to introducing point-wise pose sensitive regularization.
The BCM-M branch can outperform the baseline by 0.8\% and 4.0\% on two datasets respectively. 
This can be credited to exploiting shape priors of the CAD model provides an ideal reference for the pose regression network, which alleviates the partiality of scene point clouds due to (self-)occlusion. 
The combination of BCM-S and BCM-M further gains an improvement of 1.1\% and 4.5\% on both datasets, indicating that these two branches provide complementary information, which further supports our claim about an ensemble of pose predictions.

\begin{table}[t]
\centering
\caption{Effects of varying size of points for generating point pairs on the YCB-Video dataset. $Z$ denotes the number of FPS points sampled on input scene points or model points.}
\resizebox{0.40\textwidth}{!}{
\begin{tabular}{c|ccccccc}
\hline
$Z$ & 2 & 4 & 10 & 20 & 50 & 100 & 200 \\
\hline
$\mathcal{T}^\text{BCM-S}$ & 94.2 & 94.8 & 95.4 & 95.5 & 95.7 & 95.7 & 95.7\\
\hline
$\mathcal{T}^\text{BCM-M}$ & 94.5 & 94.9 & 95.4 & 95.6 & 95.6 & 95.6 & 95.6\\
\hline
\end{tabular}}
\label{tab:ablation_pair}\vspace{-0.3cm}
\end{table}

\paragraph{Evaluation on Size of Point-pairs.}
We evaluate pose predictions via point-pair matching by taking an average of $\mathcal{T}^\text{BCM-S}$ and $\mathcal{T}^\text{BCM-M}$ as the output pose respectively with varying size $Z$ of points, and report results in Table \ref{tab:ablation_pair}.
As our method only relies on simple geometric properties of two local points for pose computation, even with sparse input points, $\mathcal{T}^\text{BCM-S}$/$\mathcal{T}^\text{BCM-M}$ can gain comparable performance to $\mathcal{T}^\text{PR}$ (see the first row of Table \ref{tab:ablation_bcm}).

\section{Conclusion}

This paper introduces a novel neural network for 6D pose estimation based on an ensemble of direct regression and locally matching pairs of oriented points under the camera and model space.
The proposed BiCo-Net can achieve robust performance on severe inter-object occlusion and systematic noises in scene point clouds, owing to our design of exploiting pose sensitive information carried by each pair of oriented points and an ensemble of redundant pose predictions.
Experiment results can verify the effectiveness of each module in our method and the state-of-the-art performance, especially on the more challenging Occlusion LineMOD. 

\section*{Acknowledgments}
This work is supported in part by the National Natural Science Foundation of China (Grant No.: 61771201, 61902131), the Program for Guangdong Introducing Innovative and Entrepreneurial Teams (Grant No.: 2017ZT07X183).

\bibliographystyle{named}
\bibliography{ijcai22}

\begin{thebibliography}{}

\bibitem[\protect\citeauthoryear{Arun \bgroup \em et al.\egroup
  }{1987}]{arun1987least}
K~Somani Arun, Thomas~S Huang, and Steven~D Blostein.
\newblock Least-squares fitting of two 3-d point sets.
\newblock {\em TPAMI}, 1987.

\bibitem[\protect\citeauthoryear{Barron and Malik}{2013}]{barron2013intrinsic}
Jonathan~T Barron and Jitendra Malik.
\newblock Intrinsic scene properties from a single rgb-d image.
\newblock In {\em CVPR}, 2013.

\bibitem[\protect\citeauthoryear{Besl and McKay}{1992}]{besl1992method}
PJ~Besl and Neil~D McKay.
\newblock A method for registration of 3-d shapes.
\newblock {\em TPAMI}, 1992.

\bibitem[\protect\citeauthoryear{Brachmann \bgroup \em et al.\egroup
  }{2014}]{brachmann2014learning}
Eric Brachmann, Alexander Krull, Frank Michel, Stefan Gumhold, Jamie Shotton,
  and Carsten Rother.
\newblock Learning 6d object pose estimation using 3d object coordinates.
\newblock In {\em ECCV}, 2014.

\bibitem[\protect\citeauthoryear{Chen \bgroup \em et al.\egroup
  }{2020}]{chen2020g2l}
Wei Chen, Xi~Jia, Hyung~Jin Chang, Jinming Duan, and Ales Leonardis.
\newblock G2l-net: Global to local network for real-time 6d pose estimation
  with embedding vector features.
\newblock In {\em CVPR}, 2020.

\bibitem[\protect\citeauthoryear{Doumanoglou \bgroup \em et al.\egroup
  }{2016}]{doumanoglou2016recovering}
Andreas Doumanoglou, Rigas Kouskouridas, Sotiris Malassiotis, and Tae-Kyun Kim.
\newblock Recovering 6d object pose and predicting next-best-view in the crowd.
\newblock In {\em CVPR}, 2016.

\bibitem[\protect\citeauthoryear{Drost \bgroup \em et al.\egroup
  }{2010}]{drost2010model}
Bertram Drost, Markus Ulrich, Nassir Navab, and Slobodan Ilic.
\newblock Model globally, match locally: Efficient and robust 3d object
  recognition.
\newblock In {\em CVPR}, 2010.

\bibitem[\protect\citeauthoryear{He \bgroup \em et al.\egroup
  }{2017}]{he2017mask}
Kaiming He, Georgia Gkioxari, Piotr Doll{\'a}r, and Ross Girshick.
\newblock Mask r-cnn.
\newblock In {\em ICCV}, 2017.

\bibitem[\protect\citeauthoryear{He \bgroup \em et al.\egroup
  }{2020}]{he2020pvn3d}
Yisheng He, Wei Sun, Haibin Huang, Jianran Liu, Haoqiang Fan, and Jian Sun.
\newblock Pvn3d: A deep point-wise 3d keypoints voting network for 6dof pose
  estimation.
\newblock In {\em CVPR}, 2020.

\bibitem[\protect\citeauthoryear{He \bgroup \em et al.\egroup
  }{2021}]{He2021FFB6DAF}
Yisheng He, Haibin Huang, Haoqiang Fan, Qifeng Chen, and Jian Sun.
\newblock Ffb6d: A full flow bidirectional fusion network for 6d pose
  estimation.
\newblock {\em CVPR}, 2021.

\bibitem[\protect\citeauthoryear{Hinterstoisser \bgroup \em et al.\egroup
  }{2011}]{hinterstoisser2011multimodal}
Stefan Hinterstoisser, Stefan Holzer, Cedric Cagniart, Slobodan Ilic, Kurt
  Konolige, Nassir Navab, and Vincent Lepetit.
\newblock Multimodal templates for real-time detection of texture-less objects
  in heavily cluttered scenes.
\newblock In {\em ICCV}, 2011.

\bibitem[\protect\citeauthoryear{Kehl \bgroup \em et al.\egroup
  }{2016}]{kehl2016deep}
Wadim Kehl, Fausto Milletari, Federico Tombari, Slobodan Ilic, and Nassir
  Navab.
\newblock Deep learning of local rgb-d patches for 3d object detection and 6d
  pose estimation.
\newblock In {\em ECCV}, 2016.

\bibitem[\protect\citeauthoryear{Kehl \bgroup \em et al.\egroup
  }{2017}]{kehl2017ssd}
Wadim Kehl, Fabian Manhardt, Federico Tombari, Slobodan Ilic, and Nassir Navab.
\newblock Ssd-6d: Making rgb-based 3d detection and 6d pose estimation great
  again.
\newblock In {\em ICCV}, 2017.

\bibitem[\protect\citeauthoryear{Lepetit \bgroup \em et al.\egroup
  }{2008}]{Lepetit2008EPnPAA}
Vincent Lepetit, Francesc Moreno-Noguer, and P.~Fua.
\newblock Epnp: An accurate o(n) solution to the pnp problem.
\newblock {\em IJCV}, 2008.

\bibitem[\protect\citeauthoryear{Li \bgroup \em et al.\egroup
  }{2018}]{li2018unified}
Chi Li, Jin Bai, and Gregory~D Hager.
\newblock A unified framework for multi-view multi-class object pose
  estimation.
\newblock In {\em ECCV}, 2018.

\bibitem[\protect\citeauthoryear{Liebelt \bgroup \em et al.\egroup
  }{2008}]{liebelt2008independent}
Joerg Liebelt, Cordelia Schmid, and Klaus Schertler.
\newblock independent object class detection using 3d feature maps.
\newblock In {\em CVPR}, 2008.

\bibitem[\protect\citeauthoryear{Oberweger \bgroup \em et al.\egroup
  }{2018}]{oberweger2018making}
Markus Oberweger, Mahdi Rad, and Vincent Lepetit.
\newblock Making deep heatmaps robust to partial occlusions for 3d object pose
  estimation.
\newblock In {\em ECCV}, 2018.

\bibitem[\protect\citeauthoryear{Park \bgroup \em et al.\egroup
  }{2019}]{park2019pix2pose}
Kiru Park, Timothy Patten, and Markus Vincze.
\newblock Pix2pose: Pixel-wise coordinate regression of objects for 6d pose
  estimation.
\newblock In {\em ICCV}, 2019.

\bibitem[\protect\citeauthoryear{Pavlakos \bgroup \em et al.\egroup
  }{2017}]{pavlakos20176}
Georgios Pavlakos, Xiaowei Zhou, Aaron Chan, Konstantinos~G Derpanis, and
  Kostas Daniilidis.
\newblock 6-dof object pose from semantic keypoints.
\newblock In {\em ICRA}, 2017.

\bibitem[\protect\citeauthoryear{Peng \bgroup \em et al.\egroup
  }{2019}]{peng2019pvnet}
Sida Peng, Yuan Liu, Qixing Huang, Xiaowei Zhou, and Hujun Bao.
\newblock Pvnet: Pixel-wise voting network for 6dof pose estimation.
\newblock In {\em CVPR}, 2019.

\bibitem[\protect\citeauthoryear{Rad and Lepetit}{2017}]{rad2017bb8}
Mahdi Rad and Vincent Lepetit.
\newblock Bb8: A scalable, accurate, robust to partial occlusion method for
  predicting the 3d poses of challenging objects without using depth.
\newblock In {\em ICCV}, 2017.

\bibitem[\protect\citeauthoryear{Song \bgroup \em et al.\egroup
  }{2020}]{song2020hybridpose}
Chen Song, Jiaru Song, and Qixing Huang.
\newblock Hybridpose: 6d object pose estimation under hybrid representations.
\newblock In {\em CVPR}, 2020.

\bibitem[\protect\citeauthoryear{Sun \bgroup \em et al.\egroup
  }{2010}]{sun2010depth}
Min Sun, Gary Bradski, Bing-Xin Xu, and Silvio Savarese.
\newblock Depth-encoded hough voting for joint object detection and shape
  recovery.
\newblock In {\em ECCV}, 2010.

\bibitem[\protect\citeauthoryear{Sundermeyer \bgroup \em et al.\egroup
  }{2018}]{sundermeyer2018implicit}
Martin Sundermeyer, Zoltan-Csaba Marton, Maximilian Durner, Manuel Brucker, and
  Rudolph Triebel.
\newblock Implicit 3d orientation learning for 6d object detection from rgb
  images.
\newblock In {\em ECCV}, 2018.

\bibitem[\protect\citeauthoryear{Tekin \bgroup \em et al.\egroup
  }{2018}]{tekin2018real}
Bugra Tekin, Sudipta~N Sinha, and Pascal Fua.
\newblock Real-time seamless single shot 6d object pose prediction.
\newblock In {\em CVPR}, 2018.

\bibitem[\protect\citeauthoryear{Wang \bgroup \em et al.\egroup
  }{2019a}]{wang2019densefusion}
Chen Wang, Danfei Xu, Yuke Zhu, Roberto Mart{\'\i}n-Mart{\'\i}n, Cewu Lu,
  Li~Fei-Fei, and Silvio Savarese.
\newblock Densefusion: 6d object pose estimation by iterative dense fusion.
\newblock In {\em CVPR}, 2019.

\bibitem[\protect\citeauthoryear{Wang \bgroup \em et al.\egroup
  }{2019b}]{wang2019normalized}
He~Wang, Srinath Sridhar, Jingwei Huang, Julien Valentin, Shuran Song, and
  Leonidas~J Guibas.
\newblock Normalized object coordinate space for category-level 6d object pose
  and size estimation.
\newblock In {\em CVPR}, 2019.

\bibitem[\protect\citeauthoryear{Xiang \bgroup \em et al.\egroup
  }{2018}]{xiang2017posecnn}
Yu~Xiang, Tanner Schmidt, Venkatraman Narayanan, and Dieter Fox.
\newblock Posecnn: A convolutional neural network for 6d object pose estimation
  in cluttered scenes.
\newblock {\em RSS}, 2018.

\bibitem[\protect\citeauthoryear{Xu \bgroup \em et al.\egroup
  }{2018}]{xu2018pointfusion}
Danfei Xu, Dragomir Anguelov, and Ashesh Jain.
\newblock Pointfusion: Deep sensor fusion for 3d bounding box estimation.
\newblock In {\em CVPR}, 2018.

\bibitem[\protect\citeauthoryear{Zhou \bgroup \em et al.\egroup
  }{2021}]{zhou2021pr}
Guangyuan Zhou, Huiqun Wang, Jiaxin Chen, and Di~Huang.
\newblock Pr-gcn: A deep graph convolutional network with point refinement for
  6d pose estimation.
\newblock In {\em ICCV}, 2021.

\end{thebibliography}

\end{document}